\documentclass[11pt,a4paper]{article}
\usepackage{authblk}
\usepackage[hyperref]{naaclhlt2019}
\usepackage{times}
\usepackage{latexsym}
\usepackage{natbib}
\usepackage{pslatex}
\usepackage[english]{babel}
\usepackage[utf8]{inputenc}
\usepackage{amsmath}
\usepackage{bm}
\usepackage{graphicx}
\usepackage{tikz}
\usepackage{xcolor}
\usepackage{tabu}
\usepackage{multirow}
\usepackage{url}
\usepackage{rotating}
\usepackage{natbib}
\usepackage{amssymb}
\usepackage{linguex}
\usepackage{tikz}
\usepackage{tikz-qtree}

\newcommand{\key}[1]{\textsc{#1}}

\usepackage{url}

\aclfinalcopy 

\title{Structural Supervision Improves Learning of Non-Local Grammatical Dependencies}

\author[1]{\textbf{Ethan Wilcox}}
\author[2]{\textbf{Peng Qian}}
\author[3]{\textbf{Richard Futrell}}
\author[4]{\textbf{Miguel Ballesteros}}
\author[2]{\textbf{Roger Levy}}

\affil[1]{Department of Linguistics, Harvard University, \tt{wilcoxeg@g.harvard.edu}}
\affil[2]{Department of Brain and Cognitive Sciences, MIT, \tt{\{pqian,rplevy\}@mit.edu}}
\affil[3]{Department of Language Science, UC Irvine, \tt{rfutrell@uci.edu}}
\affil[4]{IBM Research, MIT-IBM Watson AI Lab \tt{miguel.ballesteros@ibm.com}}

\date{}

\begin{document}

\setlength{\Exlabelwidth}{0.7em}
\setlength{\Exlabelsep}{0.7em}
\setlength{\SubExleftmargin}{1.3em}
\setlength{\Extopsep}{2pt}

\maketitle
\begin{abstract}

State-of-the-art LSTM language models trained on large corpora learn sequential contingencies in impressive detail and have been shown to acquire a number of non-local grammatical dependencies with some success.  Here we investigate whether supervision with hierarchical structure enhances learning of a range of grammatical dependencies, a question that has previously been addressed only for subject-verb agreement.  Using controlled experimental methods from psycholinguistics, we compare the performance of word-based LSTM models versus two models that represent hierarchical structure and deploy it in left-to-right processing: Recurrent Neural Network Grammars (RNNGs) \citep{dyer2016rnng} and a incrementalized version of the Parsing-as-Language-Modeling configuration from \citet{charniak2016parsing}. Models are tested on a diverse range of configurations for two classes of non-local grammatical dependencies in English---\textit{Negative Polarity} licensing and \textit{Filler--Gap Dependencies}. Using the same training data across models, we find that structurally-supervised models outperform the LSTM, with the RNNG demonstrating best results on both types of grammatical dependencies and even learning many of the \textit{Island Constraints} on the filler--gap dependency.  Structural supervision thus provides data efficiency advantages over purely string-based training of neural language models in acquiring human-like generalizations about non-local grammatical dependencies.

\end{abstract}

\section{Introduction}
\label{sec:introduction}

Long Short-Term Memory Recurrent Neural Networks (LSTMs) \cite{hochreiter1997long} have achieved state of the art language modeling performance \cite{jozefowicz2016exploring} and have been shown to indirectly learn a number of non-local grammatical dependencies, such as subject-verb number agreement and filler-gap licensing \citep{linzen2016assessing, wilcox2018rnn}, although they fail to learn others, such as Negative Polarity Item and anaphoric pronoun licensing \citep{marvin2018targeted, futrell2018rnns}. LSTMs, however, require  large amounts of training data and remain relatively uninterpretable. One model that attempts to address both these issues is the Recurrent Neural Network Grammar \citep{dyer2016rnng}. RNNGs are generative models, which represent hierarchical syntactic structure and use neural control to deploy it in left-to-right processing. They can achieve state-of-the-art broad-coverage scores on language modeling and phrase structure parsing tasks, learn Noun Phrase headedness \citep{kuncoro2016recurrent}, and outperform linear models at learning subject-verb number agreement \citep{kuncoro2018lstms}.

In this work, we comparatively evaluate LSTMs, RNNGs and a third model trained using syntactic supervision---similar to the Parsing-as-Language-Modeling configuration from \citet{charniak2016parsing}---by conducting side-by-side tests on two novel English grammatical dependencies, deploying methodology from psycholinguistics. In this paradigm, the language models are fed with hand-crafted sentences, designed to draw out behavior that belies whether they have learned the underlying syntactic dependency. For example, \citet{linzen2016assessing} and \citet{kuncoro2018lstms} assessed how well neural language models were able to learn subject-verb number agreement by feeding the prefix \textit{The keys to the cabinet...} If the model assigns a relatively higher probability to the grammatical plural verb \textit{are} than the ungrammatical singular \textit{is} it can be said to have learned the agreement dependency. Here, we investigate two non-local dependencies that remain untested for RNNGs: Negative Polarity Item (NPI) licensing is the dependency between a negative licensor---such as \textit{not} or \textit{none}---and a Negative Polarity Item such as \textit{any} or \textit{ever}. The filler--gap dependency is the dependency between a filler---such as \textit{who} or \textit{what}---and a gap, which is an empty syntactic position. Both dependencies have been shown to be learnable by LSTMs trained on large amounts of data \citep{wilcox2018rnn, marvin2018targeted}. Here, we investigate whether, after controlling for size of the training data, explicit hierarchical representation results in learning advantages.

\section{Methodology}
\label{sec:methodology}

\subsection{Neural Language Models}

{\bf Recurrent Neural Network LMs} 
model a sentence in a purely sequential basis, without explicitly representing the latent syntactic structure. We use the LSTM architecture in \citet{hochreiter1997long}, deploying a 2-layer LSTM language model with hidden layer size 256, input embedding size 256, and dropout rate 0.3. We refer to this model as the ``LSTM" model in the following sections.

\noindent {\bf Recurrent Neural Network Grammars} \cite{dyer2016rnng} 
predict joint probability of a sentence as well as its syntactic parse. RNNGs contain three sub-components, all of which are LSTMS: the \textit{neural stack}, which keeps track of the current parse, the \textit{output buffer}, which keeps track of previously-seen terminals and the \textit{history of actions}. At each timestep the model can take three different actions: {\sc nt}, which introduces a non-terminal symbol---such as a {\sc VP} or {\sc NP}---onto the stack; {\sc shift}, which places a terminal symbol onto the top of the stack, or {\sc reduce}. {\sc Reduce} pops terminal symbols (words) off the stack until a non-terminal phrasal boundary is encountered; it then combines the terminals into a single representation via a bidirectional-LSTM and pushes the newly-reduced constituent back onto the stack. By reducing potentially unbounded constituents within the neural stack, the RNNG is able to create structural adjacency between co-dependent words that may be linearly distal. Following \newcite{dyer2016rnng}, we use 2-layer LSTMs with 256 hidden layer size for the stack-LSTM, action LSTM, and terminal LSTM, and dropout rate 0.3.

\noindent {\bf ActionLSTM}: It is the combination of the neural stack and the {\sc reduce} function that may give the RNNG an advantage over purely sequential models (such as LSTMS) or models that deploy syntactic supervision without explicit notions of compositionality. In order to assess the gains from explicitly modeling compositionality, we compare the previous two models against an incrementalized version of the Parsing-as-Language-Modeling configuration presented in \citet{charniak2016parsing}. In this model, we strip an RNNG of its \textit{neural stack} and \textit{output buffer}, and train it to jointly predict the action sequence of a parse tree as well as the upcoming word. The action space of the model contains a set of non-terminal nodes ({\sc nt}), terminal generations ({\sc gen}), as well as a ({\sc reduce}) action, which functions only as a generic phrasal boundary marker. The model was trained using embedding size 256, dropout 0.3, and was able to achieve a parsing F1 score of 92.81 on the PTB, which is only marginally better than the performance of the original architecture on the same test set, as reported in \citet{kuncoro2016recurrent}. We will refer to this model as the ``ActionLSTM" model in the following sections.

All three models are trained on the training-set portion of the English Penn Treebank standardly used in the parsing literature (PTB; sections 2-21), which consists of about \textbf{950,000 tokens} of English language news-wire text \cite{marcus1993building}. The RNNG and Action models get supervision from syntactic annotation--crucially, only constituent boundaries and major syntactic categories, with functional tags and empty categories stripped away---whereas the LSTM language model only uses the sequences of terminal words. We train the models until performance converges on the held-out PTB development-set data.


\subsection{Psycholinguistic Assessment Paradigm}
\label{sec:psycholinguistic-assessment-paradigm}
\subsubsection {Surprisal}

The \textbf{surprisal}, or negative log-conditional probability, $S(x_i)$ of a sentence's $i^{th}$ word $x_i$, tells us how strongly $x_i$  is expected in context and is also known to correlate with human processing difficulty \citep{smith2013effect,hale2001probabilistic,levy2008expectation}.  For sentences out of context, surprisal is:
\begin{equation*}
S(x_i) = -\log p(x_i|x_1 \dots x_{i-1}) 
\end{equation*}
We investigate a model's knowledge of a grammatical dependency, which is the co-variance between an upstream \textit{licensor} and a downstream \textit{licensee}, by measuring the effect that an upstream licensor has on the surprisal of a downstream licensee. The idea is that grammatical licensors should set up an expectation for the licensee thus reducing its surprisal compared to minimal pairs in which the licensor is absent. We derive the word surprisal from the LSTM language model by directly computing the negative log value of the predicted conditional probability $p(x_i|x_1 \dots x_{i-1})$ from the softmax layer. 

Following the method in \newcite{hale2018finding} for estimating word surprisals from RNNG, we use word-synchronous beam search \cite{stern2017effective} to find a set of most likely incremental parses and sum their forward probabilities to approximate $P(x_1, \dots x_i)$ and $P(x_1, \dots x_{i-1})$ for computing the surprisal.  
We set the action beam size to 100 and word beam size to 10.  We ensured that the correct incremental RNNG parses were present on the beam immediately before and throughout the material over which surprisal was calculated through manual spot inspection; the correct parse was almost always at the top of the beam.

\subsubsection{Wh-Licensing Interaction}
Unlike NPI, licensing, the filler---gap dependency is the covariance between a piece of extant material, a filler, and a piece of \textit{absent} material, a gap. Here, we employ the methodology from \citet{wilcox2018rnn}, which introduces the \textbf{Wh-Licensing Interaction}. To compute the wh-licensing interaction for a sentence, \citet{wilcox2018rnn} construct four variants, given in \ref{ex:licensing-interaction}, that exhibit the four possible combinations of fillers and gaps for a specific syntactic position. The underscores are for presentational purposes only and were not included in experimental materials.

\ex. \label{ex:licensing-interaction} 
\small
\a. I know that the lion devoured the gazelle at sunrise. \textsc{\small[-Filler -Gap]} \label{ex:nofiller-nogap}
\b. *I know what the lion devoured the gazelle at sunrise. \textsc{\small[+Filler -Gap]} \label{ex:filler-nogap}
\c. *I know that the lion devoured \_\_ at sunrise. \textsc{\small[-Filler +Gap]}\label{ex:filler-gap}
\d. I know what the lion devoured \_\_ at sunrise. \textsc{\small[+Filler +Gap]}\label{ex:nofiller-gap}

If a filler sets up an expectation for a gap, then filled syntactic positions should be more surprising in the context of a filler than in a minimally-different, non-filler variants. We measure this expectation by calculating the difference of surprisal between \ref{ex:filler-nogap} and \ref{ex:nofiller-nogap}. Similarly, if gaps require fillers to be licensed, transitions from  transitive verbs to adjunct clauses that skip an obligatory argument should be less surprising in the context of a filler than in minimally-different, non-filler variants. We measure this expectation by computing the difference in surprisal between \ref{ex:filler-gap} and \ref{ex:nofiller-gap}. Because the filler--gap dependency is a two-way interaction, the wh-licensing interaction consists of the difference of these two differences, which is given in \ref{fig:licensing-equation}.

\ex. \label{fig:licensing-equation}
$(S\ref{ex:filler-nogap} - S\ref{ex:nofiller-nogap}) - (S\ref{ex:filler-gap} - S\ref{ex:nofiller-gap})$ 

For basic filler---gap dependencies, we expect the presence of a filler to set up a global expectation for a gap, thus we measure the summed licensing interaction across the entire embedded clause, which we expect to be significantly above zero if the model is learning the dependency. Our experimental materials include only vocabulary items within the PTB, avoiding the need for Out of Vocabulary handling. We determine statistical significance using a mixed-effects linear regression model, using sum-coded conditions \citep{baayen2008mixed}. For within-model comparison we use surprisal as the dependent variable and experimental conditions as predictors; for between-model comparison, we use wh-licensing interaction as the dependent variable with model type and experimental conditions as predictors. All figures depict  by-item means, with error bars representing 95\% confidence intervals, computed by subtracting out the within-item means from each condition as advocated by \citet{masson2003using}. The strength of a wh-licensing interaction can be interpreted as either its mean size in bits, or as its mean size normalized by its standard deviation across items.  The latter is Cohen's $d$, rooted in signal-detection theory (\citealp{cohen:1977statistical-power-analysis}); because all our experiments involve similar number of items, it is roughly proportional to the size of wh-interaction relative to the size of the associated confidence interval.\footnote{All of our experiments were pre-registered online at \texttt{http://aspredicted.org/blind.php?x=\{xd9cw9, 3xv2du, jd384m, cy6zp6, 2hk4gf, zt73qt, f9pk9f, ab9f3h, yt6pi4\}}}


\section{Negative Polarity Item Licensing}
\label{sec:npi-licensing}

In English, Negative Polarity Items (NPIs), such as \textit{any}, \textit{ever} must be in the \key{scope} of a negative \key{licensor} such as \textit{no}, \textit{none}, or \textit{not} \citep{fauconnier:1975polarity,ladusaw:1979polarity}. Crucially, the scope of a licensor is characterized structurally, not in purely linear terms; for present purposes, a sufficient approximation is that an NPI is in the proper scope of a licensor if it is c-commanded by it.  Thus while \textit{ever} in~\ref{ex:npi-no-the} and~\ref{ex:npi-no-no} is grammatical because it is licensed by \textit{no} in the main-clause subject, \textit{ever} is ungrammatical in \ref{ex:npi-the-no} despite the linearly preceding \textit{no}, because inside a subject-modifying relative clause is not a valid position for an NPI licensor; we call this a \key{distractor} position.

\ex. \label{ex:npi-licensing}
\small
\a. *\textbf{The} senator that supported the measure has ever found any support from her constituents. \label{ex:npi-the-the}
\b. \textbf{No} senator that supported the measure has ever found any support from her constituents. \label{ex:npi-no-the}
\c. * \textbf{The} senator that supported \textbf{no} measure has ever found any support from her constituents. \label{ex:npi-distractor}\label{ex:npi-the-no}
\d. \textbf{No} senator that supported no measure has ever found any support from her constituents.\label{ex:npi-no-no}

Learning of NPI licensing conditions by LSTM language models trained on large corpora has previously been investigated by \citet{marvin2018targeted} and \citet{futrell2018rnns}.  \citeauthor{futrell2018rnns} found that the language models of both \citet{gulordava2018colorless} and \citet{jozefowicz2016exploring} (hereafter called `Large Data LSTMs') learned a contingency between licensors and NPIs: the NPIs in examples like~\ref{ex:npi-licensing} were lower-surprisal when linearly preceded by negative licensors.  However, both papers reported that these models failed to constrain the contingency along the correct structural lines: negative NPI surprisal was decreased at least as much by a preceding negative distractor as by a negative licensor.

\begin{figure}[t]
\begin{minipage}{0.5\textwidth}
\centering
\includegraphics[width=0.63\textwidth]{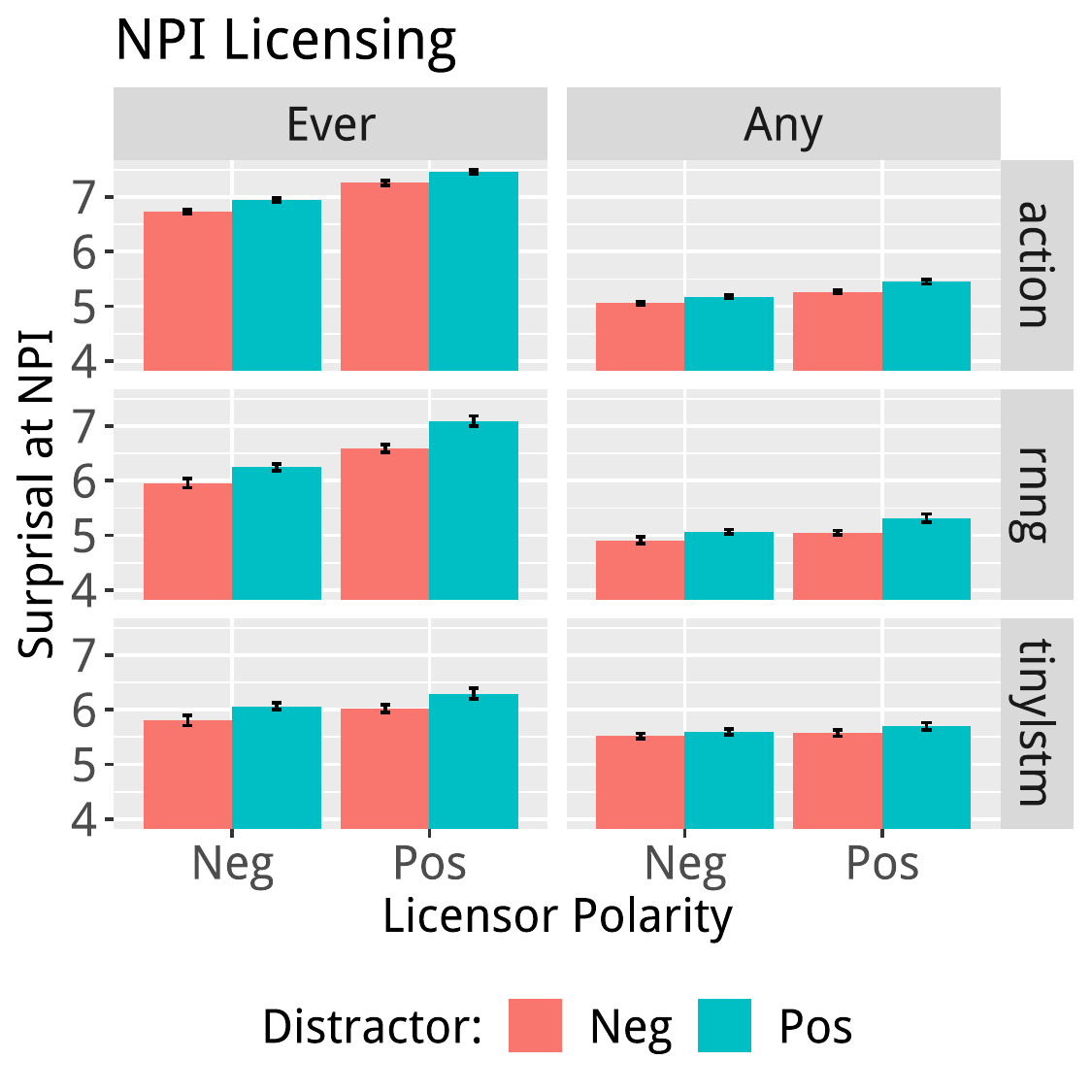}
\includegraphics[width=0.34\textwidth]{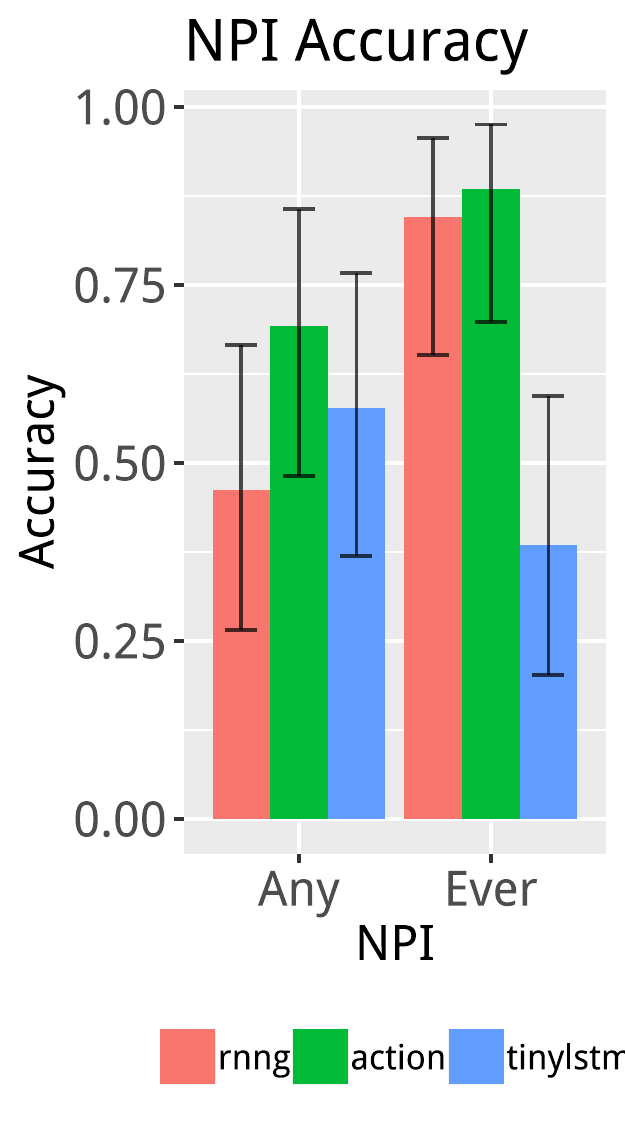}
\end{minipage}
\caption{NPI Licensing at left: Y-axis shows surprisal at the NPI, x-axis indicates polarity of the c-commanding licensor, and color indicates distractor polarity. Licensing accuracy at right: Y-axis shows classification accuracy,  x-axis indicates the NPI tested, and color indicates the model. Error bars represent 95\% binomial confidence intervals.}
\label{fig:npi-results}
\end{figure}

Syntactic supervision might plausibly facilitate learning of NPI licensing conditions.  We tested this following the method of \citet{futrell2018rnns}, constructing 27 items on the design of in \ref{ex:npi-licensing}, with two variants: one included \textit{ever} and omitting \textit{any}, and one including \textit{any} and omitting \textit{ever}.  Figure~\ref{fig:npi-results}, left panel, shows the results.  For the RNNG and the ActionLSTM, negative licensors and distractors alike reduced surprisal of both NPIs ($p<0.05$ for the RNNG, $p<0.001$ for the ActionLSTM).  For the LSTM, negative licensors and distractors alike reduced surprisal of \textit{ever} (both $p<0.01$), but not \textit{any}.  This may seem surprising as \textit{any} is considerably more frequent than \textit{ever} (123 vs.\ 727 instances in the training data), but \textit{any}'s non-NPI uses (e.g., \emph{I will eat anything fried}) may complicate its learning.

From Figure~\ref{fig:npi-results} it is also apparent that the RNNG and ActionLSTM show signs of stronger NPI licensing effects from negation in the licensor position than in the distractor position, at least for \textit{ever}. To quantify this, we follow \citet{marvin2018targeted} in computing item-mean classification accuracies, with classification being considered correct if the NPI is assigned higher probability in context for \ref{ex:npi-no-the} than for \ref{ex:npi-the-no}.  Results are shown in Figure \ref{fig:npi-results}, right panel. No model is significantly above chance for \textit{any}, but for \textit{ever} the syntactically supervised models perform much better: The RNNG reaches 85\% performance, and the ActionLSTM 88\%, both significantly above chance ($p<0.001$ by binomial test for each), and are not significantly different from each other, but both better than the LSTM ($p<0.01$ for the RNNG/LSTM; $p<0.001$  for the ActionLSTM/LSTM by Fisher's exact test).  To our knowledge this is the first demonstration of a language model learning the licensing conditions for an NPI without direct supervision.

Overall, we find that syntactic supervision facilitates the contingency of NPIs on a negative licensor in context, but is not sufficient for clean generalization of the structural conditions on NPI licensing with the training dataset used here.

\section{Filler--Gap Dependencies}

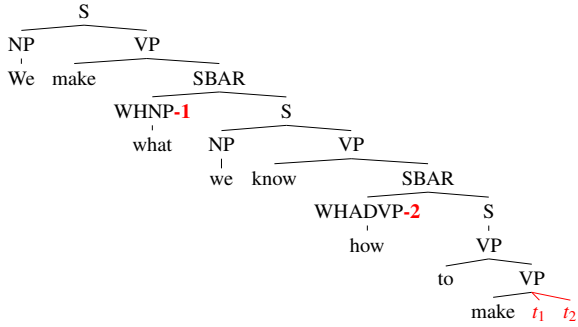
\begin{figure}
    \centering
    \scalebox{0.7}{
\begin{tikzpicture}
\tikzset{level distance=18pt}
\tikzset{sibling distance=2pt}
    \Tree [.S [.NP We ] [.VP make [.SBAR [.{WHNP\textcolor{red}{\textbf{-1}}} \node(what){what}; ] [.S [.NP we ] [.VP know [.SBAR [.{WHADVP\textcolor{red}{\textbf{-2}}} \node(how){how}; ] [.S [.VP to [.\node(lowestVP){VP}; make \edge[color=red]; \textcolor{red}{$t_1$} \edge[color=red]; \textcolor{red}{$t_2$} ] ] ] ] ] ] ] ] ] 
\end{tikzpicture}
}
\caption{Example of filler--gap dependency representation in the Penn Treebank.  Non-local dependency annotation indicated in bold, red font is stripped from the training data, so that the RNNG must learn about the filler-gap dependency purely through neural generalization.}
    \label{fig:filler-gap-annotation-in-PTB}
\end{figure}

The dependency between a \textsc{filler}, which is a wh-word such as \textit{who} or \textit{what}, and a \textsc{gap}, which is an empty syntactic position, is characterized by a number of properties, some of which were tested for large data LSTMs by \citet{wilcox2018rnn}.  Here we investigate the effect of syntactic supervision on filler--gap dependency learning.  Syntactic annotation of the dependency itself is stripped from the training data (Figure~\ref{fig:filler-gap-annotation-in-PTB}), so syntactic supervision can play only an indirect facilitatory role for the models' neural learning mechanisms.

\subsection{Flexibility of Gap Position}\label{sec:flexibility}

\begin{table}
    \footnotesize
    \centering
    \begin{tabular}{|l|c|c|c|}
        \hline
         Location of Gap & All Fillers & `Who' & `What' \\
        \hline
        All Positions & 13907 & 1888 & 660 \\
        Subject Position & 6632 & 1510 & 236 \\
        Object Position & 2080 & 12 & 332 \\
        Indirect Object Position & 57 & 0 & 6 \\
        \hline
    \end{tabular}
    \caption{Filler---Gap Dependency Statistics for the Penn Treebank training data (used for both models).}
    \label{tab:fgdep}
\end{table}

The filler--gap dependency is flexible: a filler can license a gap in any of a number of syntactic positions, including the argument positions of subject, object, and indirect object, as illustrated in \ref{ex:flexability}, as well as in other positions (e.g.\ the adjunct position for \emph{how} in Figure~\ref{fig:filler-gap-annotation-in-PTB}). 

\ex. \label{ex:flexability}
\small
\a. I know who \_\_ introduced the accountant to the guests after lunch.
\b. I know who the CEO introduced \_\_ to the guests after lunch.
\c. I know who the CEO introduced the accountant to \_\_ after lunch.

\noindent
These gap positions differ in frequency, however (Table \ref{tab:fgdep}): the majority (63.1\%) are in some argument structure position, of which the vast majority (75.6\%) are subject position (mostly subject-extracted relative clauses), 23.7\% are object position, and 0.7\% are indirect object position.  

\begin{figure*}[!ht]
\centering
\includegraphics[width=0.25\textwidth]{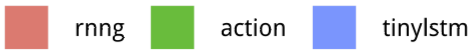}

\begin{minipage}{0.45\textwidth}
\scriptsize
\centering
\includegraphics[width=\textwidth]{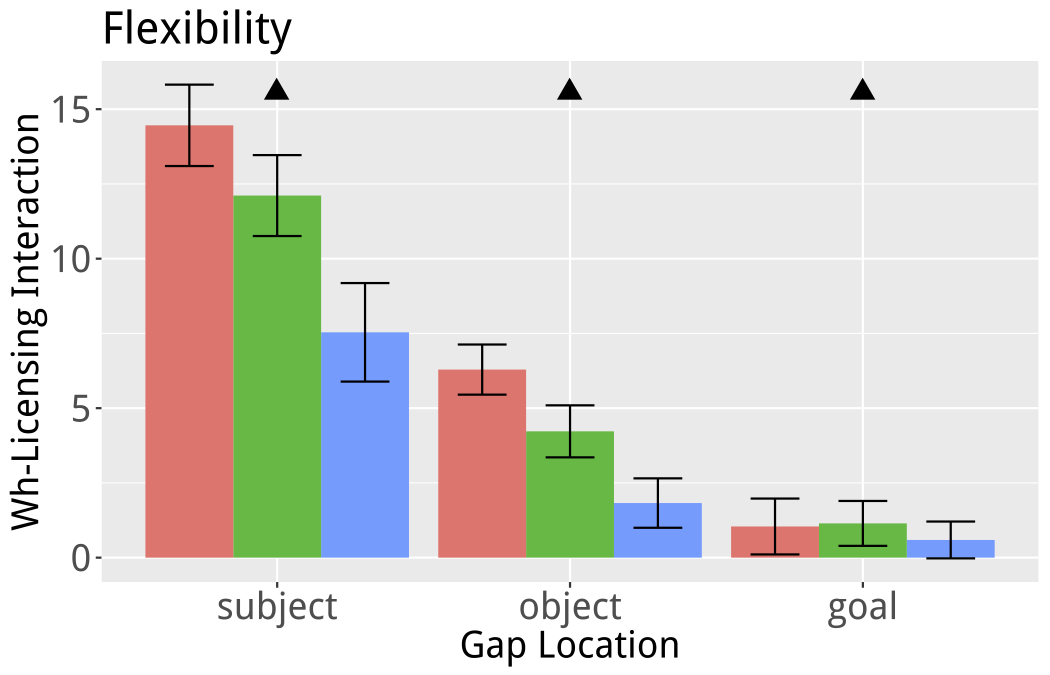}
\end{minipage}
\hspace{4pt}
\begin{minipage}{0.45\textwidth}
\scriptsize
\centering
\includegraphics[width=\textwidth]{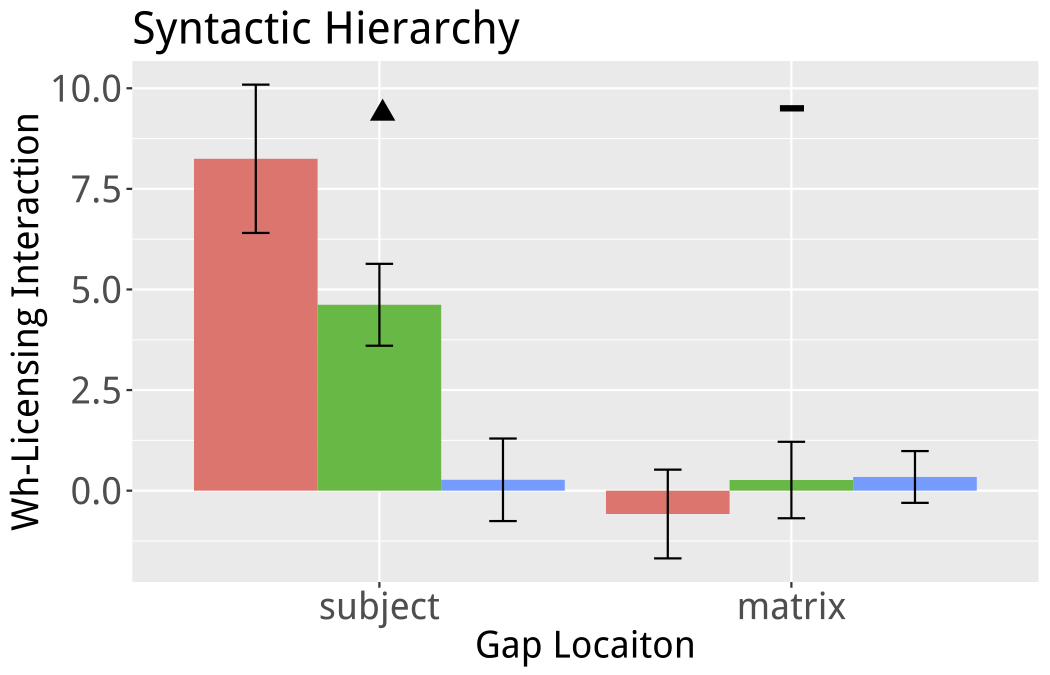}
\end{minipage}
\vspace*{2cm}
\begin{minipage}{0.45\textwidth}
\scriptsize
\centering
\includegraphics[width=\textwidth]{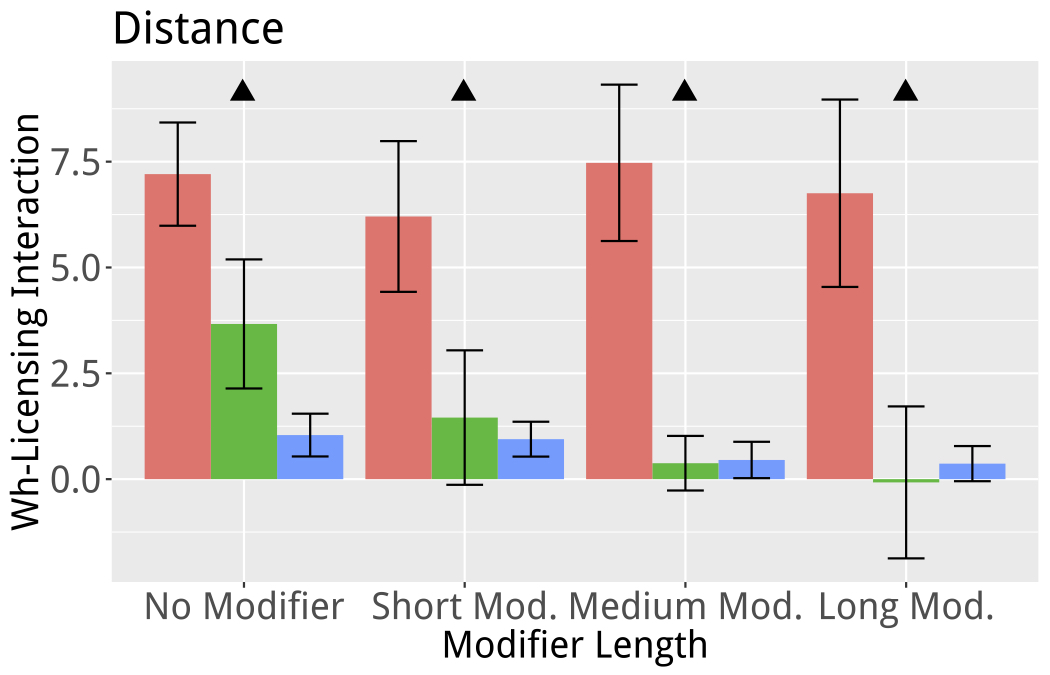}
\end{minipage}
\hspace{4pt}
\begin{minipage}{0.45\textwidth}
\scriptsize
\centering
\includegraphics[width=\textwidth]{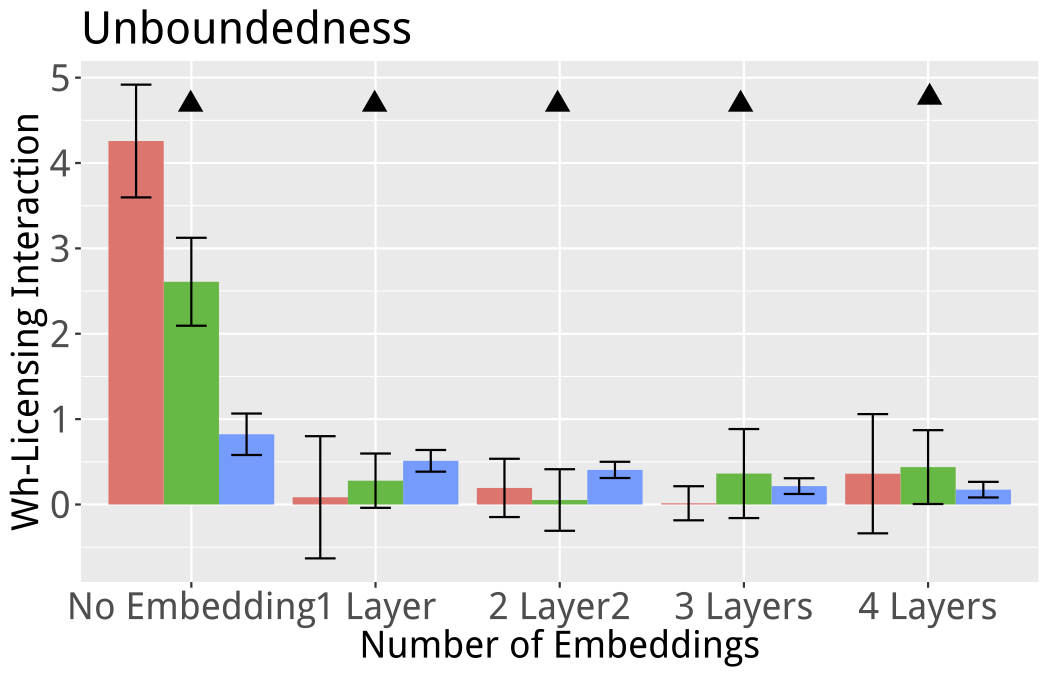}
\end{minipage}
\vspace{-2cm}
\caption{Model results for the basic properties of filler--gap licensing. ``$\blacktriangle$" indicates grammatical conditions in which models should display strong wh-licensing interaction, ``\textbf{--}" indicates ungrammatical conditions in which models should display reduced wh-licensing interaction. The RNNG model significantly outperforms the LSTM model in 8/13 grammatical cases; the ActionLSTM model outperforms the LSTM model in 5/13 cases; and the RNNG outperforms the ActionLSTM model in 6/13 cases where strong licensing is expected.}
\label{fig:fillergap-results}
\end{figure*}

Using the wh-interaction measure described in Section~\ref{sec:psycholinguistic-assessment-paradigm}, \citet{wilcox2018rnn} showed that large-data LSTMs learn filler--gap dependencies for all three argument positions, with the size of the wh-interaction generally largest for subject gaps and smallest for indirect-object gaps.  Table~\ref{tab:fgdep} suggests that this gradation may reflect frequency of learning signal, with the dependency being learned more robustly the more frequent the extraction type.  We applied the same method, adapting \citeauthor{wilcox2018rnn}'s items to the smaller training dataset.  The results can be seen in the upper-left panel of Figure~\ref{fig:fillergap-results}.  

All three models learn the filler-gap dependency for subject and object positions, and there is suggestive but inconclusive evidence for learning in the rare indirect object position. We see stronger dependency learning for more frequent gap types, as was found for large data LSTMs, and the supervised models show a much stronger wh-licensing effect than the LSTM.

\subsection{Syntactic Hierarchy}
\label{sec:syntactic-hierarchy}

As with NPIs, the filler--gap dependency is subject to a number of hierarchical, structural constraints. The most basic of these constraints is that the filler must be ``above" the gap in the appropriate structural sense (to a first approximation, the filler must $c$-command the gap, though see e.g.\  \citealp{pollard1994headdriven} for qualifications).  Hence \emph{who} in \ref{ex:hierarchy-good} is a legitimate extraction from the relative clause, but \ref{ex:hierarchy-bad} is ungrammatical as the gap is in the matrix clause, above the filler.
\ex. \label{ex:hierarchy}
\small
\a. The policeman who the criminal shot \_\_ with his gun shocked the jury during the trial. \label{ex:hierarchy-good}
\b. *The policeman who the criminal shot the politician with his gun shocked \_\_ during the trial. \label{ex:hierarchy-bad}

A model that properly generalizes this constraint on the filler--gap dependency should \emph{not} show a wh-interaction for cases like \ref{ex:hierarchy-bad}: an undischarged \emph{who} filler should not make the matrix-clause gap particularly more expected.  As far as we are aware, no prior work has investigated this property of the filler--gap dependency in language models; we do so here. Because the context in \ref{ex:hierarchy} does not allow for an immediate \emph{that} clause initiation for the \textsc{--Filler} condition as in \ref{ex:licensing-interaction}, we instantiate this condition by contrasting the \textsc{+Filler,+Gap} condition of \ref{ex:hierarchy-bad} with the variants in~\ref{ex:hierarchy-bad-other-conditions}, where the \emph{who} filler is immediately discharged as the RC verb's extracted subject:
\ex. \label{ex:hierarchy-bad-other-conditions}
\small
\a. *The policeman who \_\_ knows that the criminal shot the politician with his gun shocked \_\_ during the trial. \textsc{-Filler,+Gap}
\b. *The policeman who the criminal shot the politician with his gun shocked the jury during the trial. \textsc{+Filler,--Gap}
\c. The policeman who \_\_ knows that the criminal shot the politician with his gun shocked the jury during the trial. \textsc{-Filler,--Gap}
\z.

We created 22 items following the templates of \ref{ex:hierarchy-good} (\emph{Subject} condition) and \ref{ex:hierarchy-bad} (\emph{Matrix} condition); results are shown in the top-right panel of \ref{fig:fillergap-results}.  The supervised models show a large wh-licensing interaction effect for a gap inside the subject-modifying relative clause---with the RNNG demonstrating more licensing interaction than the ActionLSTM---and neither model inappropriately generalizes this licensing effect to a matrix-clause gap.  The LSTM shows no wh-licensing effects in either position, suggesting that syntactic supervision facilitates appropriately generalized filler-gap dependencies for subject-modifying relative clauses.\footnote{Results for the Larger Data LSTM models for the Hierarchy and Unboundedness experiments presented here can be found in the appendix.}


\subsection{Robustness to Intervening Material}

For a model that learns human-like syntactic generalizations and maintains accurate phrase-like representations throughout a string, filler--gap dependencies should be robust to linearly intervening material that does not change the tree-structural relationship between the filler and the gap. \citet{wilcox2018rnn} found that the large-data RNNs described earlier exhibit a robust wh-interaction of this type, by introducing an optional postnominal modifier between filler and gap to sentence templates like \ref{ex:modification}, with no modification \ref{ex:mod-none}, short (3--5 word) modifiers \ref{ex:mod-short}, medium (6--8 word) modifiers \ref{ex:mod-medium}, and long (8--12 word) modifiers \ref{ex:mod-long}.

\ex. \label{ex:modification}
\small
\a. I know what your friend gave \_\_ to Alex last weekend. \label{ex:mod-none}
\b. I know what your friend in the hat gave \_\_ to Alex last weekend. \label{ex:mod-short}
\c. I know what your friend who you ate lunch with yesterday gave \_\_ to Alex last weekend. \label{ex:mod-medium}
\d. I know what your friend who recently took you on a walking tour of the city gave \_\_ to Alex last weekend. \label{ex:mod-long}

\noindent
We adapted their materials for the small training dataset and tested our three models; results are shown in \ref{fig:fillergap-results}, bottom-left panel. The RNNG shows a robust licensing interaction that does not diminish with additional intervening material (all $d>1.3$). The LSTM shows smaller wh-licensing interactions across the board; these are still substantial in the \textit{No Modifier} and \textit{Short Modifier} conditions ($d=0.88, d=0.98$, respectively), but are smaller in the \textit{Medium Modifier} and \textit{Long Modifier} conditions ($d=0.45, d=0.37$ respectively), suggesting less robustness to intervening material. The ActionLSTM shows strong interactions in the \textit{No Modifier} condition ($d=1.02$), but weak interaction once any modifying material is introduced ($d<0.4$ in all other conditions). This result is significant, as it indicates that RNNG is able to leverage the structural locality afforded by the neural stack to maintain robust gap expectancy.

\subsection{Unboundedness} 
\label{sec:unboundedness}

For humans, filler--gap dependencies are not only robust to linearly intervening material that does not change their tree-structural relationship, they can be \key{structurally non-local} as well, propagating through intervening syntactic structures (subject to constraints examined in Section~\ref{sec:island-constraints}).  For example, a filler can be extracted from multiply-nested complement clauses as in \ref{ex:unboundedness-five}:

\ex. \label{ex:unboundedness}
\small
\a. I know who your aunt insulted \_\_ at the party.
\b. I know who the chauffeur said [$_\text{S}$ the hostess believed [$_\text{S}$ the butler reported [$_\text{S}$ her friend thinks your aunt insulted \_\_ at the party.]]] \label{ex:unboundedness-five}

Humans show sensitivity to a single layer of sentential embedding when processing filler---gap dependencies in an offline `complexity rating' task \cite{phillips2005erp}. This may due to the relative frequency of single versus doubly-embedded filler---gap dependencies. In our training data there were 13,907 examples of filler---gap dependencies, however only 758 examples that spanned two layers of sentential embedding and 19 that spanned three layers. There were no instances of filler---gap dependencies spanning over more than three sentential embeddings, as in \ref{ex:unboundedness-five}.  

The unboundedness of filler--gap dependencies has not previously been tested for contemporary language models.  To do this, we constructed 22 test items like \ref{ex:unboundedness}, varying embedding depth within-item between zero, one, two, three, and four levels, and measured the resulting licensing interactions. The results are in Figure \ref{fig:fillergap-results}, bottom-right panel.  No model's filler--gap dependency is perfectly robust to clausal embedding.  The LSTM's wh-licensing interaction starts out small and diminishes with embedding depth.  The RNNG and ActionLSTM show strong wh-licensing interaction in the unembedded condition but no significant wh-licensing interaction after even a single layer of embedding.  Since these experimental materials are new, we also tested the large-data LSTMs on them, which exhibited much larger and more robust filler--dependency effects (Appendix~B). Hence the syntactic supervision explored here is not sufficient to guarantee that learned filler--gap dependencies can be structurally unbounded.


\section{Island Constraints}
\label{sec:island-constraints}

\begin{figure}
    \centering
\begin{tikzpicture}
\tikzset{level distance=18pt}
\tikzset{sibling distance=2pt}
\Tree [.{} \edge[roof]; {$\alpha$} [.{} 
[.\node(filler){filler}; ]
[.{} \edge[roof]; {$\beta$} [.\textbf{X}
\edge[roof]; 
{$\gamma$} \node(gap){\_\_}; \edge[roof]; {$\delta$}
] \edge[roof]; {$\zeta$}  ] \edge[roof]; {$\nu$} ] {$\theta$} ]
\draw[style=dashed] (filler.south) .. controls +(south:1.5) .. node {\large$\times$} (gap.south);
    \end{tikzpicture}
\caption{Anatomy of an island constraint. If node \textbf{X} is an island, then a filler outside \textbf{X} cannot associate with a gap inside \textbf{X}.  For our analyses, successful learning of an island constraint implies that we should \emph{not} see a wh-licensing interaction at the first part of the material $\delta$ immediately following the potential gap site.}
    \label{fig:island-constraint-anatomy}
\end{figure}
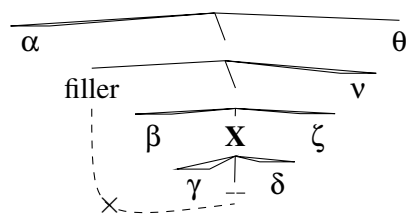

A crucial exception to the flexibility and unboundedness of filler--gap dependencies is that \key{Island Constraints} prevent association of a filler and a gap through certain types of syntactic nodes, illustrated in Figure~\ref{fig:island-constraint-anatomy} \citep{ross1967constraints}. Contemporary theories variously attribute island effects to grammatical rules, incremental processing considerations, or discourse-structural factors \citep{ambridge2008island, hofmeister2010cognitive, sprouse2013experimental}. 
In our setting, a language model is sensitive to an island constraint if it \emph{fails} to show a wh-licensing interaction between a filler and a gap that cross an island.
\citet{wilcox2018rnn} found evidence that large-data LSTMs are sensitive to some island constraints (although see \citet{chowdhury2018rnn} for a contrasting view), but not to others. Here we investigate whether LSTMs would learn these from smaller training datasets, and if an RNNG's syntactic supervision provides a learning advantage for island constraints. In this section we measure the wh-licensing interaction in the material immediately following the potential gap site, which is guaranteed to implicate the model's (lack of) expectation for a gap inside the island, rather than throughout the entire embedded clause, which also implicates filler-driven expectations after the end of the island.

\subsection{Adjunct Islands}
\label{sec:adjunct-islands}

Adjunct clauses block the filler--gap dependency. \citet{wilcox2018rnn} found evidence that large-data LSTMs are sensitive to adjunct islands, as evidenced by attenuated and often fully eliminated wh-licensing interactions for materials like \ref{ex:adj-back}--\ref{ex:adj-front} relative to \ref{ex:adj-obj} below.  (In this and the subsequent subsections, the post-gap material used for wh-interaction computation is in \textbf{bold}.)

{\small
\ex. \label{ex:islands:adjunct}
\small
\a. The director discovered what the robbers stole \_\_ \textbf{last night}. \textsc{[Object]}\label{ex:adj-obj}
\b. *The director discovered what the security guard slept while the robbers stole \_\_ \textbf{last night}. \textsc{[Adj-back]}\label{ex:adj-back}
\c. *The director discovered what, while the robbers stole \_\_ \textbf{last night}, the security guard slept. \textsc{[Adj-front]}\label{ex:adj-front}
\z.

}
%
	%
We adapted these materials; results are in Figure~\ref{fig:island-results}, upper-left panel. The RNNG shows a strong licensing interaction in the baseline main-clause object extraction position, but no licensing interaction for a gap in an adjunct either at the back or front of the main clause. Because RNNGs failed our test for unboundedness of filler--gap dependency, however (Section~\ref{sec:unboundedness}), this result is inconclusive as to whether anything corresponding to an island constraint is learned. The LSTM and the ActionLSTM show no sign of filler--gap dependency attenuation from adjunct islands, in contrast to previous findings using the LSTM architecture on much larger training datasets.


\subsection{Wh Islands}

\begin{figure*}[!ht]
\centering
\includegraphics[width=0.25\textwidth]{images/legend.png}

\begin{minipage}{0.45\textwidth}
\centering
\includegraphics[width=\textwidth]{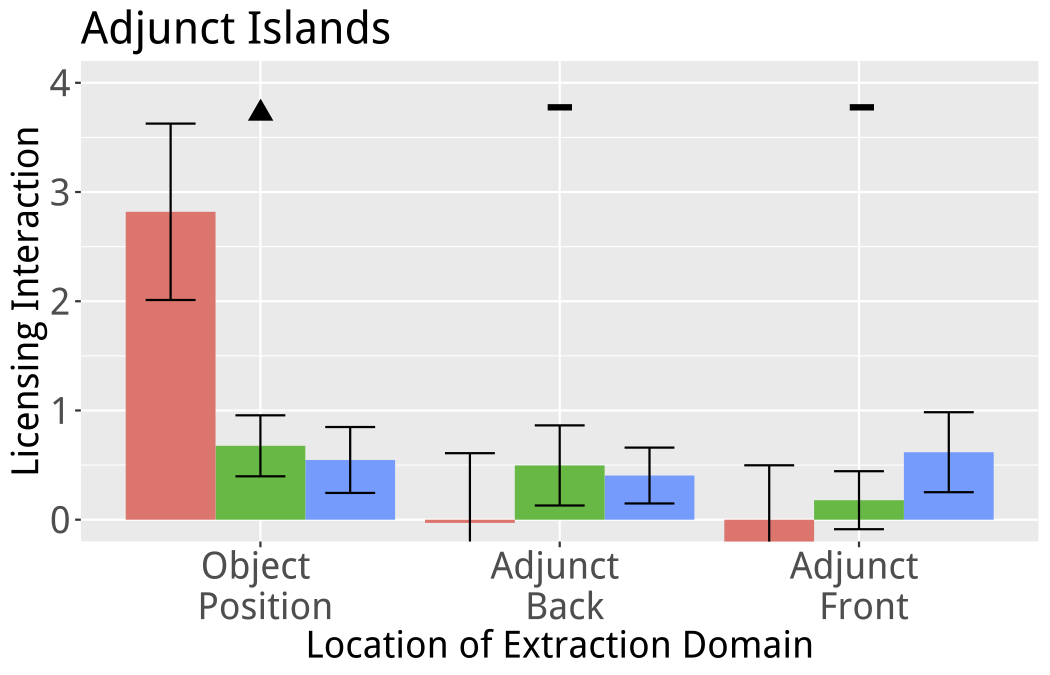}
\end{minipage}
\hspace{4pt}
\begin{minipage}{0.45\textwidth}
\centering
\includegraphics[width=\textwidth]{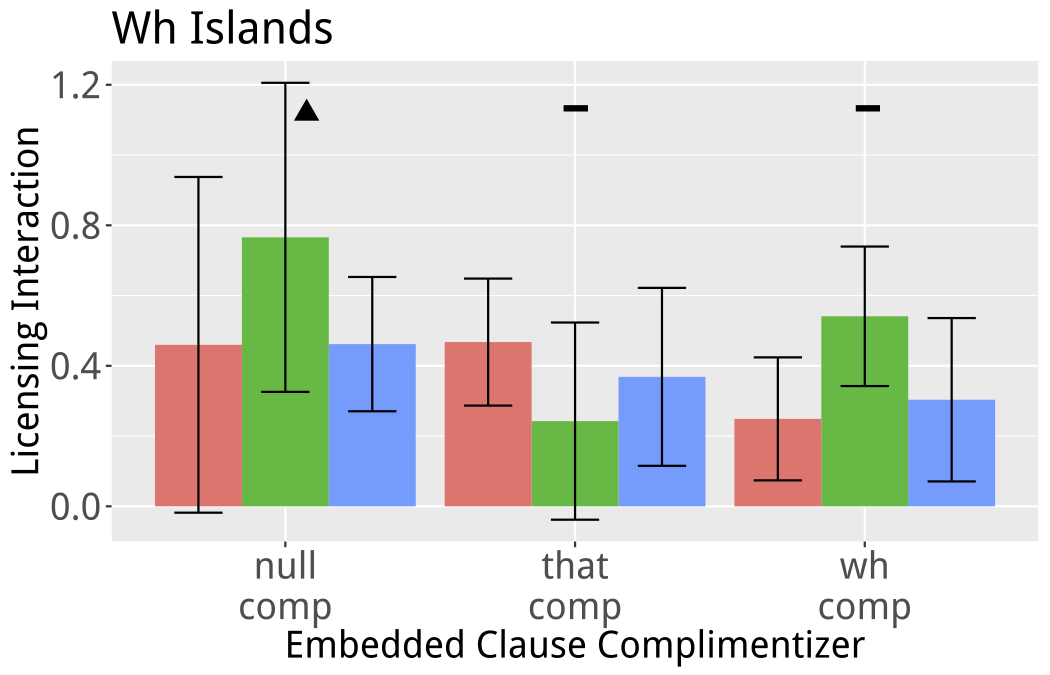}
\end{minipage}
\begin{minipage}{0.45\textwidth}
\centering
\includegraphics[width=\textwidth]{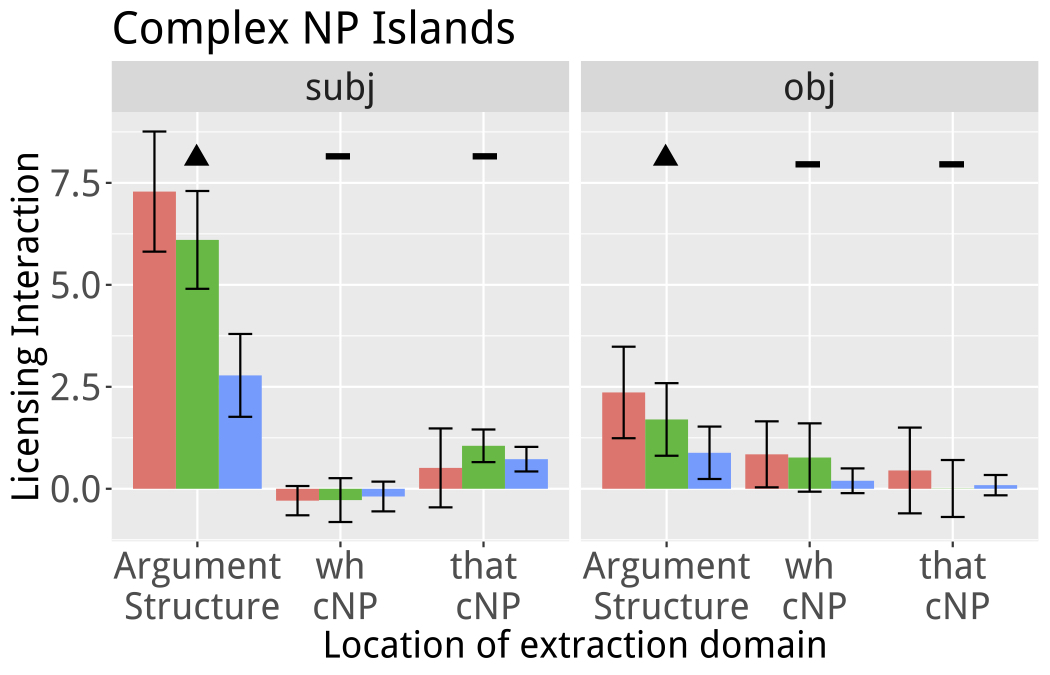}
\end{minipage}
\hspace{4pt}
\begin{minipage}{0.45\textwidth}
\centering
\includegraphics[width=\textwidth]{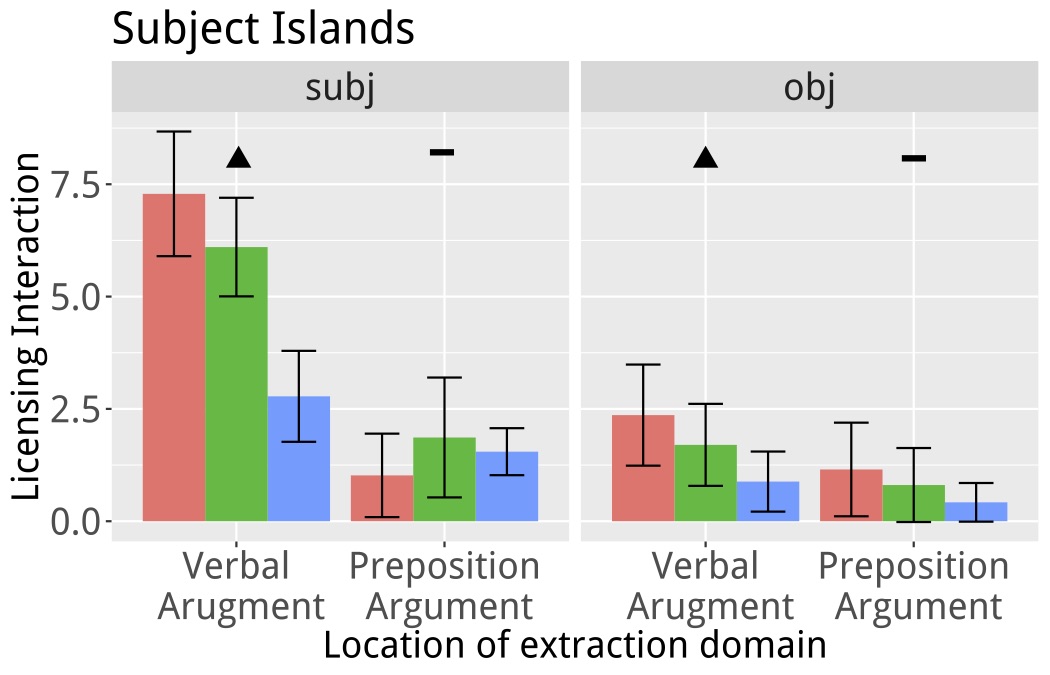}
\end{minipage}
\caption{Model results for Syntactic Islands.  ``$\blacktriangle$" indicates grammatical conditions in which models should display strong wh-licensing interaction, ``\textbf{--}" indicates ungrammatical conditions in which models should display reduced wh-licensing interaction.}
\label{fig:island-results}
\end{figure*}

Embedded sentences introduced by \textit{wh-} words are also islands; hence, \ref{ex:wh-comp} is anomalous but \ref{ex:null-comp} and \ref{ex:that-comp} are not.  

{\small
\ex. \label{ex:islands-wh}
\a. 
    I know what the guide said the lion devoured \_\_ \textbf{yesterday}. [\textsc{Null Comp}]\label{ex:null-comp}
\b. I know what the guide said that the lion devoured \_\_ \textbf{yesterday}. [\textsc{That Comp}]\label{ex:that-comp}
\c. *I know what the guide said whether the lion devoured \_\_ \textbf{yesterday}. [\textsc{Wh- Comp}]\label{ex:wh-comp} 
\z.
	
}

\noindent
\citet{wilcox2018rnn} found that the large-data LSTMs learned this island constraint: the wh-licensing interaction was eliminated or severely attenuated for the \textsc{Wh-Comp}lementizer variant but not for the other variants.
Results for our three models are in Figure~\ref{fig:island-results}, top-right panel.  These materials paint a slightly more optimistic picture than the results of Section~\ref{sec:unboundedness} for the RNNG's ability to propagate a gap expectation from a filler down one level of clausal embedding.  However, no models show an appreciable attenuation of the wh-licensing interaction in the \textsc{Wh- Comp} condition that would suggest an island constraint-like  generalization.

\subsection{Complex Noun-Phrase Islands}

Extractions from within clauses dominated by a lexical head noun are unacceptable; this is the Complex Noun Phrase Constraint. For example, \ref{ex:cnpc-wh} and \ref{ex:cnpc-that} are unacceptable object extractions compared with \ref{ex:cnpc-obj}; the same acceptability pattern holds for subject extractions.

{
\small
\ex. \label{ex:islands-cnpc}
\a. I know what the collector bought \_\_ \textbf{last week}. [\textsc{Argument} extraction] \label{ex:cnpc-obj}
\b. *I know what the collector bought the painting which depicted \_\_ \textbf{last week}. [\textsc{Wh- Complex NP}]\label{ex:cnpc-wh}
\c. *I know what the collector bought the painting that depicted \_\_ \textbf{last week}. [\textsc{That- Complex NP}]\label{ex:cnpc-that}
\z. 

}

\noindent
\citet{wilcox2018rnn} found that large-data LSTM behavior reflected this island constraint, with attenuated wh-licensing interactions for complex NPs like \ref{ex:cnpc-wh}--\ref{ex:cnpc-that} and for analogous complex NPs involving subject extractions. Our results for adaptations of their materials are shown in Figure~\ref{fig:island-results}, bottom-left panel. All three models show attenuated wh-licensing interactions inside complex NPs in subject position, with the licensing interaction in the grammatical {\sc argument structure} position greatest for the RNNG and ActionLSTM. This may be taken as an indication of Complex NP Constraint-like learning, but is inconclusive due to the models' general failure to propagate gap expectations into embedded clauses (Section~\ref{sec:unboundedness}).

\subsection{Subject Islands}
\label{sec:subject-islands}
Prepositional phrases attaching to subjects are islands: this is the Subject Constraint, and accounts for the unacceptability of \ref{ex:subj-preparg} compared to \ref{ex:subj-verbalarg} \citep{huang1998logical}.

{
\small
\ex. \label{ex:islands-subject}
\a. I know what the collector bought \_\_ \textbf{yesterday}. [\textsc{Obj Verbal-Arg}]
\b. I know what the collector bought a painting of \_\_ \textbf{yesterday}. [\textsc{Obj Prep-Arg}]
\c. I know what \_\_ \textbf{sold} for a high price at auction. [\textsc{Subj Verbal-Arg}]\label{ex:subj-verbalarg}
\d. *I know what a painting of \_\_ \textbf{sold} for a high price at auction. [\textsc{Subj Prep-Arg}]\label{ex:subj-preparg}
\z.

}

\noindent
\citet{wilcox2018rnn} found that the wh-licensing interactions of large-data LSTMs fail to distinguish between subject-modifying PPs, which cannot be extracted from, and object-modifying PPs, which can.  Our results for adaptations of their materials can be seen in Figure~\ref{fig:island-results}, bottom right panel. The syntactically supervised models show a significant decrease between the verbal argument and prepositional argument conditions in subject position ($p<0.001$ for RNNG; $p<0.01$ for ActionLSTM), and no significant difference between the two conditions in object position (however, note that the licensing in object position is significantly less than the licensing in the grammatical, \textit{Verbal Argument Subject} position, following the pattern in \ref{sec:flexibility}).  LSTMs fare worse, showing a clear wh-licensing interaction for subject-modifying PPs, which should be islands, and no wh-licensing interaction for object-modifying PPs.

\section{Conclusion}

In this paper we have argued that structural supervision provides advantages over purely string-based training of neural language models in acquiring more human-like generalizations about non-local grammatical dependencies. We have also demonstrated how the neural compositionality of the RNNG architecture can provide even further advantages, especially at maintaining expectations into structurally-local but linearly distant material. We compared RNNG, ActionLSTM and LSTM models using recently developed controlled experimental materials, and developed additional experimental materials to further test several characteristics of grammatical dependency learning for neural language models (Sections~\ref{sec:syntactic-hierarchy}, \ref{sec:unboundedness}). We found advantages for syntactic supervision in learning conditions for \textbf{Negative Polarity Item licensing} and a majority of tests involving \textbf{filler--gap dependencies}, showing particularly strong wh-licensing effects in tree-structurally-local contexts. On basic filler---gap dependency properties the RNNG significantly outperformed the LSTM in 8/13 and the ActionLSTM outperformed the LSTM on 5/13 cases where strong licensing interaction was expected. While the RNNG, and to some extent the ActionLSTM, exhibited more humanlike behavior than the LSTM for a number of \textbf{Island Constraints}, the tests were inconclusive due to  the models' failure to propagate gap expectation into embedded clauses: island-like behavior may merely be sensitivity to general syntactic complexity, not the highly-specific syntactic arrangements that constitute the family of island constructions. Thus, major-category supervision does not provide enough information for the neural component to learn fully robust and human-like filler---gap dependencies from 1-million words alone. However, for some dependencies tested (i.e. NPIs) structural supervision on 1 million words provides better outcomes than even large-data LSTMs. Scaling the gains derived from structural supervision is a challenge for data-scarce NLP and is the basis for future work.

\section*{Acknowledgments}
This work was supported by the MIT-IBM Watson AI Lab.


\bibliographystyle{acl_natbib}
\bibliography{naaclhlt2019,everything}

\appendix
\noindent\textbf{\large \underline{Appendix}}\vspace{5pt}\\
We present results for two large data LSTM models on novel experiments described in the paper. The two models tested here are the `BIG LSTM + CNN Inputs' from \citet{jozefowicz2016exploring} (the `google' model) and the highest-performing model presented in the supplementary materials of \citet{gulordava2018colorless} (the `Gulordava' model). Both models where shown in \citep{wilcox2018rnn} to represent filler---gap dependencies and some island constraints.
\vspace{-2pt}
\section{Syntactic Hierarchy}
\begin{figure}
\begin{minipage}{0.45\textwidth}
\includegraphics[width=\textwidth]{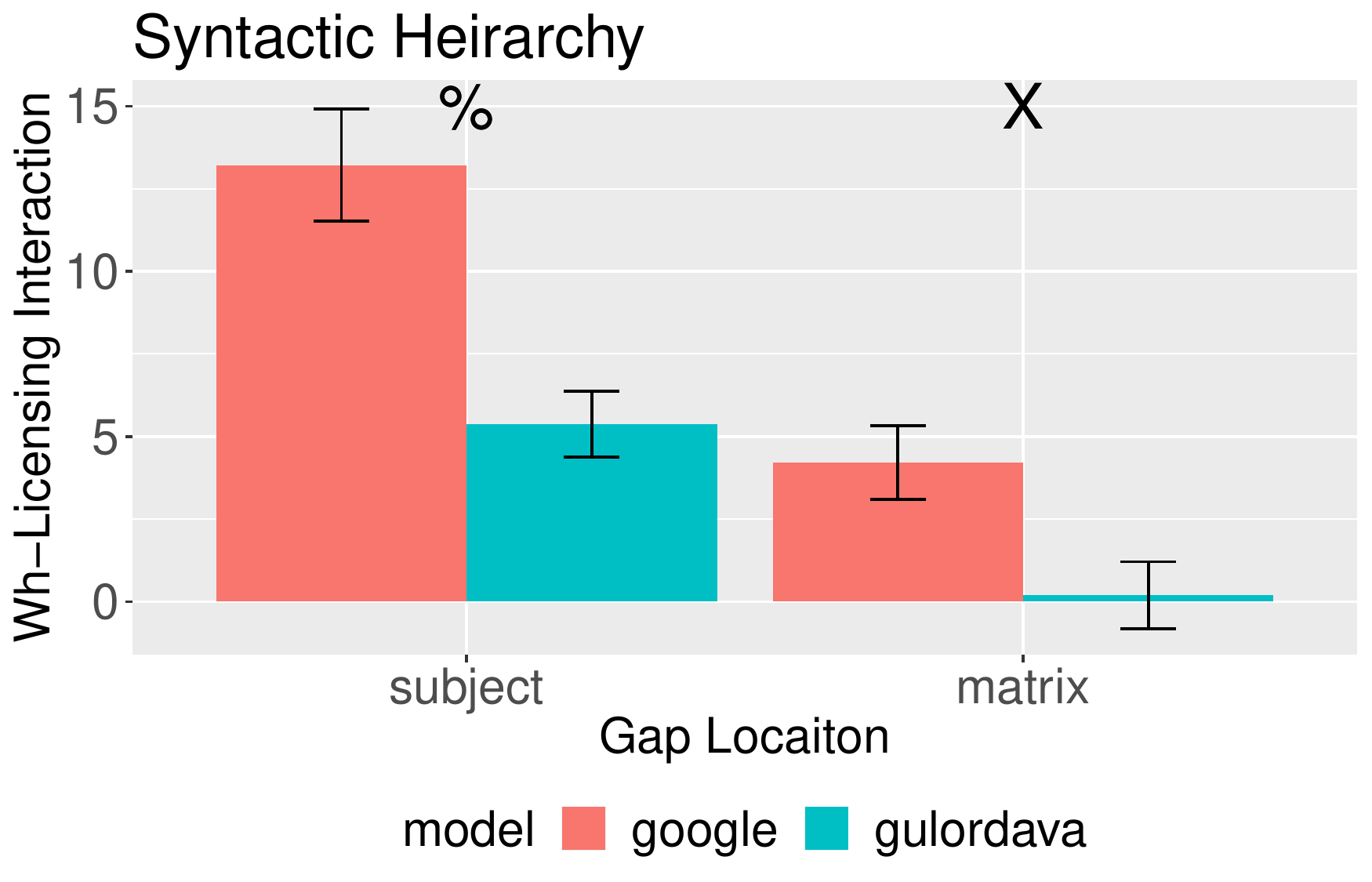}
\end{minipage}
\caption{Syntactic Hierarchy. \%s indicate conditions where we would expect a strong wh-licensing interaction, Xs indicate conditions where we expect a reduction in wh-licensing interaction.}
\label{fig:heirarchy-largelstms}
\end{figure}

\vspace{-2pt}
We tested the two large LSTM models using our stimuli from the syntactic hierarchy experiment and measured the wh-licensing interaction across the entire embedded clause. The results of this experiment can be seen in Figure \ref{fig:heirarchy-largelstms}. Both models show significant licensing interaction in the grammatical \textit{Subject} condition ($p<0.001$), and a significant reduction in licensing interaction between the \textit{Subject} and \textit{Matrix} conditions ($p<0.001$ in both models). Additionally, there is a significant licensing interaction in the \textit{Matrix} condition for the Google model, but not so for the Gulordava model.
\vspace{-2pt}
\section{Unboundedness}
\begin{figure}
\begin{minipage}{0.45\textwidth}
\includegraphics[width=\textwidth]{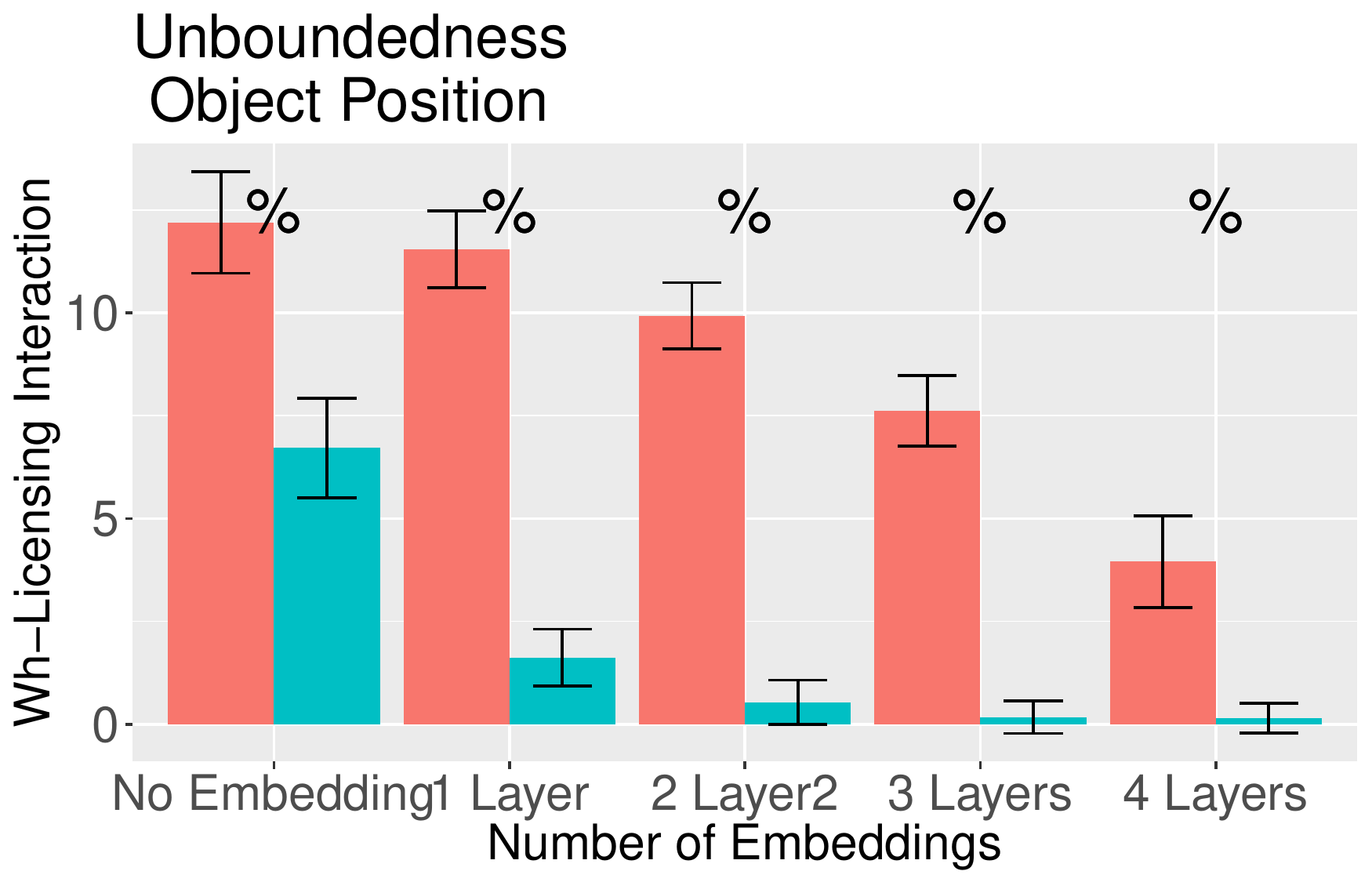}
\includegraphics[width=\textwidth]{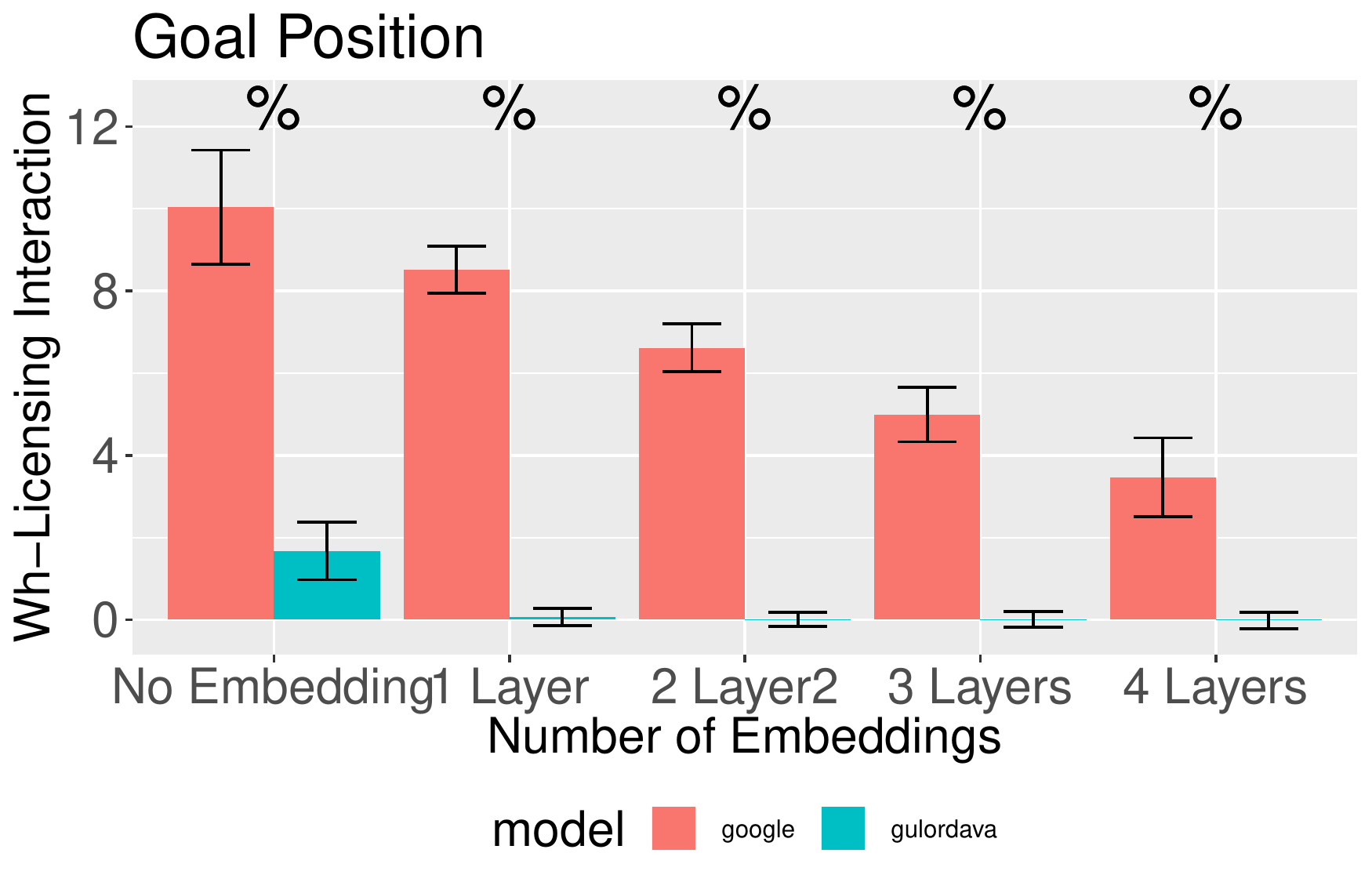}
\end{minipage}
\caption{Unboundedness, \%s indicate conditions where we would expect a strong wh-licensing interaction, Xs indicate conditions where we expect a reduction in wh-licensing interaction.}
\label{fig:unboundedness-largelstms}
\end{figure}

\vspace{-2pt}
We tested the two large LSTM models from \citet{wilcox2018rnn} following the stimuli from our unboundedness experiment, with two variants, one that included gaps in \textit{Object} position and one that included gaps in indirect object or \textit{Goal} position. The results can be seen in Figure \ref{fig:unboundedness-largelstms}. For the Google model in \textit{Object} position, we find a significant reduction of wh-licensing interaction across more than three layers of embedding ($p<0.001$). For the Gulordava model, we find a significant reduction in wh-licensing interaction after only one layer of embedding ($p<0.001$). In the \textit{Goal} position: For the Google model, we find a significant reduction in licensing interaction after two layers of embedding ($p<0.05$ for 2 layers, $p<0.001$ for 3-4 layers). For the Gulordava model, we find no significant licensing interaction after one layer of embedding. These results indicate the larger LSTMs are able to thread gap expectation through embedded clauses.

\end{document}